\RequirePackage[hyphens]{url}

\documentclass[11pt,a4paper]{article}
\usepackage[hyperref]{acl2021}
\usepackage{times}
\usepackage{latexsym}
\usepackage{nicefrac}

\PassOptionsToPackage{hyphens}{url}\usepackage{hyperref}
\usepackage{natbib}
\usepackage{enumitem}
\setlist{itemsep=0pt}

\usepackage{tikz}

\usepackage{enumitem}

\usepackage{microtype}

\usepackage{amsmath}
\usepackage{bbm}

\aclfinalcopy % Uncomment this line for the final submission
\title{Mordecai 3: A Neural Geoparser and Event Geocoder}
  
\author{Andrew Halterman \\ Department of Political Science \\ Michigan State University \\  \href{mailto:ahalterman0@gmail.com}{ahalterman0@gmail.com}}

\date{}

\begin{document}

\thispagestyle{plain}
\pagestyle{plain}

\maketitle
\begin{abstract}
Mordecai3 is a new end-to-end text geoparser and event geolocation system. The system performs toponym resolution using a new neural ranking model to resolve a place name extracted from a document to its entry in the Geonames gazetteer. It also performs event geocoding, the process of linking events reported in text with the place names where they are reported to occur, using an off-the-shelf question-answering model. The toponym resolution model is trained on a diverse set of existing training data, along with several thousand newly annotated examples. The paper describes the model, its training process, and performance comparisons with existing geoparsers. The system is available as an open source Python library, Mordecai 3, and replaces an earlier geoparser, Mordecai v2, one of the most widely used text geoparsers \citep{halterman2017mordecai}.
\end{abstract}

\section{Introduction}

Text geoparsing, the process of identifying place names in text and resolving them to their entry in a geographic gazetteer, is a key step in making text data useful for researchers, especially in social science. This paper introduces a new Python library for geoparsing documents. Specifically, it uses spaCy's named entity recognition system to identify place names in text and queries the Geonames gazetteer \citep{wick2011geonames} in a custom Elasticsearch index for candidate matches. The core novelty of the library is a new model that uses a new neural similarity model to select the best match from the candidate locations. The model is trained on a large set of existing and newly annotated text with correct geolocations.

The geoparser also performs \textit{event geolocation} \citep{halterman2019linking}, the process of linking an event in text to the location where it is reported to occur. To do so, it uses an off-the-shelf question-answering model to select which location, among potentially several in the text, corresponds to the location where the event was reported to occur.

The library performs well on both new and existing geoparsing datasets, with an country-level accuracy of 94.2\% and an exact match accuracy of 82.8\%. \footnote{The library is available here: \url{https://github.com/ahalterman/mordecai3/}.}

\begin{figure*}[h!]
\center
	\includegraphics[width=0.9\textwidth]{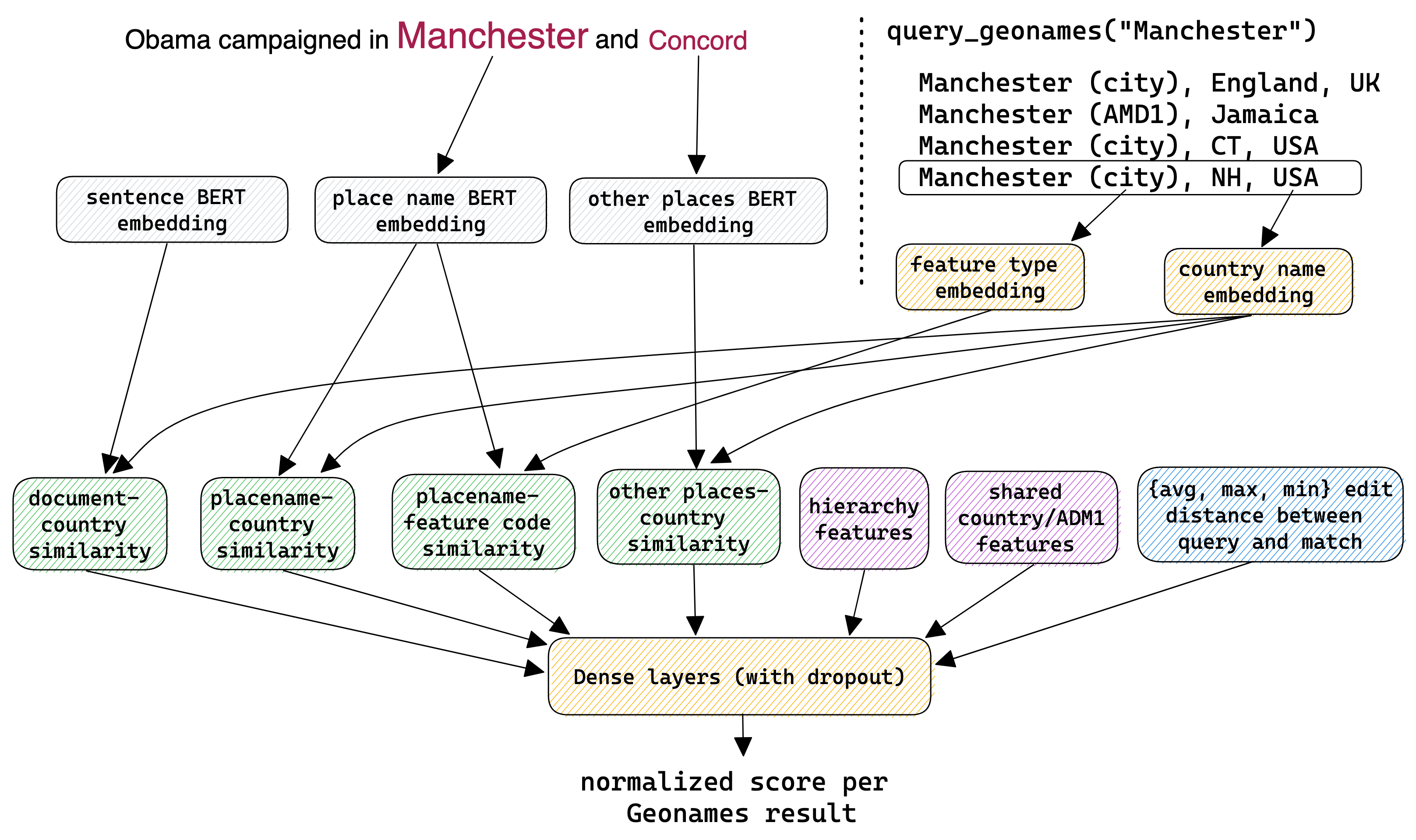}
	\caption{High-level overview of Mordecai 3's ranking model. Neural similarity comparisons are in green. Pink values are calculated using other place names in the text. Blue values are produced by comparing the original query to the Geonames gazetteer results.}
	\label{fig:model}
\end{figure*}

\section{Previous Work}

The existing work on geoparsing has identified a set of useful heuristics, including population as a strong baseline \citep{leidner2008toponym}, a ``spatial minimality" that selects candidate locations that are closest in space \citep{leidner2017georeferencing}, a ``one meaning per document" heuristic that all mentions of a place name refer to the same location, and the number of alternative names in the gazetteer as a proxy for importance \citep{halterman2017mordecai, karimzadeh2019geotxt}. Other work has used static word embeddings to infer the likely country for a place name and other heuristics \citep{halterman2017mordecai}. Current state-of-the-art models employ string similarity between a place name and candidate entries in the gazeetter, along with contextual information from the broader document \citep{wang2019dm_nlp}. This model draws on much of this work, employing string similarity methods, the length of alternative names in Geonames, and the document's contextual information in the form of transformer-based word embeddings.
	
\section{Data}

As training data, I draw on several existing datasets of resolved toponyms:

\begin{itemize}
	\item GeoWebNews \citep{gritta2019pragmatic}, 6,612 toponyms, filtered down to 2,401 that report a Geonames ID.
	\item TR News (1,275 toponyms with Geonames IDs) \citep{kamalloo2018coherent}.
	\item Local Global Corpus (LGL), which has includes primarily US local news stories \citep{lieberman2010geotagging}
\end{itemize}

Two other common geoparsing datasets, \textbf{GeoVirus} \citep{gritta2018melbourne} and \textbf{WikiTor} \citep{gritta2017missing}, 
provide links to Wikipedia pages but not Geonames, and thus are not used.

It also uses the following newly collected data:

\begin{itemize}
	\item Newly annotated data from news wire and newspaper stories (N=1,671)
	\item Synthetic data generated with a rule-based system that focuses on hierarchical place names (for example, \textit{capital} of \textit{country}, \textit{city} in \textit{state}, etc.) (N=944)
	\item New Wikipedia-derived data. I obtain training data by exporting articles from certain Wikipedia categories (battles, protests, etc), and then:
	\begin{itemize}
		\item Running NER on the article. If a named entity lines up with an internal Wiki page link, then:
		\item Follow the link to the Wikipedia page. Get that page's Wikidata ID, and then:
		\item See if the Wikidata page has a Geonames ID
	\end{itemize}
\end{itemize}

I create a random train/validation set of data from all sources to fit and evaluate the model. I also add impossible cases, where the correct entry is removed from the candidate locations, to simulate real-world situations where the location may not be present in the search results.

\begin{table*}[!ht]
    \centering
    \begin{tabular}{lllllllll}
    \hline
        \textbf{Dataset} & \textbf{Eval N} & \textbf{Exact} & \textbf{Mean Error}  & \textbf{Median}   & \textbf{Correct} & \textbf{Correct}  & \textbf{Correct}  & \textbf{Acc@}  \\ 
        & & \textbf{Match} & \textbf{(km)} & \textbf{Err. (km)} & \textbf{Country} & \textbf{Type} & \textbf{ADM1} & \textbf{161km}\\
        \hline
Training & 6673 & 83.1\% & 388.3 & 0.0 & 92.2\% & 86.4\% & 86.9\% &  88.8 \\
prodigy & 500 & 90.4\% & 65.4 & 0.0 & 97.0\% & 91.0\% & 95.6\% &  98.5 \\
TR & 273 & 81.0\% & 733.9 & 0.0 & 89.7\% & 85.0\% & 82.8\%  & 82.8\\
GWN & 477 & 91.2\% & 93.4 & 0.0 & 95.8\% & 91.6\% & 93.9\% &  96.0\\
Synth & 300 & 96.7\% & 133.6 & 0.0 & 97.7\% & 99.3\% & 97.0\% &  97.3\\
Wiki & 356 & 80.6\% & 180.4 & 0.0 & 93.0\% & 82.0\% & 89.9\% & 95.0 \\
        \hline
    \end{tabular}
    \caption{Accuracy figures for Mordecai 3 on new and existing datasets.}
    \label{tbl:acc}
\end{table*}

\section{Model}

The geoparsing process consists of the following steps:

\begin{enumerate}
	\item Identify all place names (toponyms) in a document using spaCy's named entity recognition model. \citep{honnibal2017spacy}.\footnote{Currently, it uses spaCy v3.4.3 and spaCy's transformer-based \texttt{en\_core\_web\_trf} v3.4.1.} \\
	Then, for each extracted place name, it
	\item Queries the Geonames gazetteer of placenames \citep{wick2011geonames} hosted in a custom Elasticsearch instance with the extracted placename to identify a set of candidate entries in the gazetteer. The query uses fuzzy search over both the primary place name in Geonames and the list of alternative names.
	\item Ranks the set of candidate entries using a neural model that uses similarity features and features derived from the place names in the gazetteer again. Figure \ref{fig:model} provides an overview.
	\item Optionally, if used alongside an event data system, identifies the location in a document where an event is most likely to have occurred, using an off-the-shelf extractive question-answering model.
\end{enumerate}

The model uses a set of similarity measures to select the best location out of the candidate locations.

First, it uses it computes several neural text similarity measures (shown in green in Figure \ref{fig:model}) to address the problem of ambiguous place names. It computes several similarities to infer the likely country for a location. Specifically, it learns a custom embedding for each country and computes the similarity between that embedding and the (average) spaCy transformer emebddings for the place name of interest, the other place names in the text, and the document as a whole. The spaCy transformer embeddings, from spaCy's fine-tuned RoBERTa model, we get ``for free" from the previous NER step. By using the embeddings, we can draw on contextual clues about the likely country discussed in a piece of text.  It does a similar process for inferring the geographic type of the place name (e.g., city, administrative area). This helps use contextual information to differentiate cities and administrative regions with the same name.
 
 It then also draws on string similarity between the query location and each of the candidate locations returned by Geonames (shown in blue in Figure \ref{fig:model}). These include the minimum and average edit distance between the query and the set of names for each candidate location. 
 
 Finally, it uses information from all locations identified in a document to help resolve each of them. For each candidate location, it identifies whether other locations in the document have a hierarchical relationship (e.g., neighborhood within city, city with administrative area) \citep{karimzadeh2019geotxt} and whether a candidate location shares a country with candidate locations for other place names in the text.
 
 Finally, all of these features are concatenated into a vector and passed through a dense layer. Each candidate location is given a $[0, 1]$ score, and these are softmaxed over all candidate locations to generate a single score. The model also has the option of selecting a null candidate, to handle situations where the place name may not have an entry in the gazetteer or where the earlier search step failed to identify the true location.

The model is similar to  DM\_NLP, the current state-of-the-art model \citep{wang2019dm_nlp}, which uses string similarity measures and incorporates document context. However, this model differs in that it does not require contextual information from Wikipedia and uses document context in the form of RoBERTa-based contextual embeddings, rather than a bag-of-words model. 

Note that the model does not fine-tune the transformer weights directly. Instead, it learns dense layers on top of the static spaCy embeddings. I do this for several reasons. First, fine-tuning the embeddings would require either a second transformer model or would risk major degradation of spaCy's NER performance as the model as the transformer weights were updated to perform well on non-NER tasks. Adding a second transformer model would greatly increase the computational cost and time needed to geoparse documents. Second, by not fine-tuning the weights, I ensure that the model is not overfit to the text and locations that are present in the training data. The model is intended to perform well on many kinds of text from all regions of the world, and learning location-specific features could degrade this performance. This decision stands in contrast to some existing approaches, for example, the CamCoder model introduced by \citet{gritta2018melbourne}, which learns that terms like ``pyramid complex'' and ``archeological site'' are predictive of Giza, Egypt. Similarly, \citet{speriosu2013text} train a classifier for each place name, using document context to predict each place name's correct geolocation. Both of these are limited in their applicability to places outside the training corpus.

In fitting the model, I experiment with several hyperparameters, including the batch size, dropout value, learning rate, and the country and feature-type embedding dimension. I also experiment with a gradient accumulation step and a multi-task output that attempts to predict the place name's country using the contextualized embeddings. The hyperparameters with the greatest improvement in accuracy were the epochs ($=15$), batch size ($=60$), dropout ($=0.3$), and the learning rate ($=0.4$).  

\section{Results and Evaluation}

\begin{table}
\begin{center}
\begin{tabular}{lll}
\hline
\textbf{Dataset} & \textbf{\% missing@50} & \textbf{\% missing@500} \\
\hline
		training  & 5.9\% & 3.8\% \\
New data      & 0.4\% & 0.0\% \\
TR           & 8.1\% & 1.1\% \\
LGL          & 5.9\% & 3.4\% \\
GWN          & 4.5\% & 2.9\% \\
Synth        & 1.0\% & 0.3\% \\
Wiki         & 5.9\% & 5.1\% \\
\hline
\end{tabular}
\end{center}
\caption{Evaluation of the query step. Percentage of queries without the correct answer in either the top 50 or 500 results from Elasticsearch.}
\label{tbl:query}
\end{table}

The first evaluation I conduct is the ability of the Elaticsearch query to correctly retrieve the correct location from the Geonames index. If the correct location is not retrieved, then the model will not be able to identify it. Table \ref{tbl:query} shows the percentage of correct locations that are not in the top 50 and 500 results, respectively, for each dataset. The correct location is located in the top 500 results in almost all cases. Some penalty is paid for restricting the query to the top 50 locations, but the speed improvements could make this a useful tradeoff in some situations.

\citet{gritta2019pragmatic} offer a detailed discussion of how to evaluate geoparsers and argue for an approach to evaluating geoparsing that focuses on three metrics: the AUC of the model, the percentage of results that are within 161 km (100 miles) of the true location, and the mean error. Table Table \ref{tbl:acc} presents mean distance and percent within 161 km, but also reports several other metrics that are important for end users. Specifically, it also reports the \textit{exact match} percentage. While \citet{gritta2019pragmatic} are correct to point out that the distance between the predicted and true location is an important factor in evaluating geoparsers, it is also true that errors of any magnitude are can cause problems in certain applications. I also report the proportion of locations that are resolved to the correct country, top-level administrative area (e.g. state or province), and to the correct feature type (e.g. settlement vs. administrative area).
\begin{table}[b]
\begin{center}
\begin{tabular}{llll}
\hline
\textbf{Geoparser} & \textbf{Mean} & \textbf{Acc@}  \\
& \textbf{Error} & \textbf{161km} \\
\hline
SpacyNLP + CamCoder & 188 & \textbf{95}  \\
SpacyNLP + Population & 210 & \textbf{95}  \\
Oracle NER + CamCoder & 232 & 94  \\
Oracle NER + Population & 250 & 94  \\
Yahoo Placemaker & 203 & 91  \\
Edinburgh  Geoparser & 338 & 91  \\
Mordecai3 & \textbf{184} &  94 \\
\hline
\end{tabular}
\end{center}
\caption{A reproduction of Table 4 from \citet{gritta2019pragmatic}, comparing geoparser's performance on the GWN corpus with values for Mordecai (not trained on GWN) added.}
\label{tbl:gritta_comparison}
\end{table}
The model performs well on most metrics and datasets. It has an average country-level accuracy of 94.2\% and an average exact match accuracy of 82.8\%. For all datasets, the median error is 0 km, but the high mean error indicates that some incorrectly geolocated places are resolved to locations that are very far away. It performs very well on the synthetic data, which samples locations and adds them to simple sentence templates. This text is not representative of real-world text, however. It performs worst on the LGL corpus, which heavily samples local US news. Many of the stories refer to ambiguous place names (e.g., the classic ``Springfield" example) and are written for local audiences with assumed knowledge of the area being described. 

Table \ref{tbl:gritta_comparison} reproduces a table from \citet{gritta2019pragmatic}, which compares several existing geoparser's performance on the GWN corpus \citep{gritta2019pragmatic}, and adds the performance for Mordecai3 trained on all data except GWN. It shows competitive performance, with the lowest mean error and an accuracy@161 km that is within one point of the best models.

Finally, I evaluate the model's ability to handle instances where the correct location is not present in the candidate locations returned by the query step. In most applied situations, incorrectly geolocating a place name is a worse error than failing to geolocate a place name that could have been geolocated. Table \ref{tbl:missing} shows the proportion of ``impossible" choices (i.e., instances where the correct location is not present in the candidate locations returned in the query step) that the neural ranking model correctly identifies as unanswerable.  The model shows wide variance across datasets, ranging from 40\% to 100\% accuracy in abstaining from picking a location. Training on more instances where the correct location has been manually removed from the candidate locations could improve the model's ability to abstain from impossible geolocations.

\begin{table}
\begin{tabular}{lll}
\hline
	\textbf{Dataset} & \textbf{Missingness correctly} & \textbf{Percentage} \\
	& \textbf{identified} & \textbf{Missing} \\
\hline
training  & 70.3\% & 10.8\% \\
New data  & 100.0\% & 7.0\% \\
TR &  66.7\% & 2.2\% \\
LGL & 40.0\% & 5.2\% \\
GWN &  54.7\% & 5.5\% \\
Synth &  97.4\% & 12.7\% \\
Wiki & 55.9\% & 9.6\% \\
\hline
\end{tabular}
\caption{The model's ability to identify when the correct place name is not present in the candidate locations.}
\label{tbl:missing}
\end{table}

\section{Conclusion and Future Work}

Two areas of future work could improve the performance of the model. First, more training data could improve the performance of the model. This data can be efficiently obtained in two ways. First, using the process I outline above, more training data can easily be obtained from Wikipedia. To ensure that the model remains general to other text, more human annotations can also be collected. Using the Geonames query portion of the library, candidate geolocations can be shown to annotators, who can select the correct one, lowering the cost of collecting more hand annotations.

Second, the model itself could be improved. As discussed above, the transformer models themselves are not fine-tuned, which limits the ability of the model to incorporate contextual clues about the correct location. Training a second model to predict the geographic coordinates of the location using only the context of the story, as \citet{radford2021regressing} suggests, could help select the correct location or better estimate when none of the candidate locations are correct and the model should return no correct geolocation at all.

\section{Acknowledgements}

This work was sponsored by the Political Instability Task Force (PITF). The PITF is funded by the Central Intelligence Agency. The views expressed in this article are the author's alone and do not represent the views of the US Government.

\bibliographystyle{chicago}
\bibliography{/Users/ahalterman/MIT/MIT.bib}

\begin{thebibliography}{}

\bibitem[\protect\citeauthoryear{Gritta, Pilehvar, and Collier}{Gritta et~al.}{2018}]{gritta2018melbourne}
Gritta, M., M.~Pilehvar, and N.~Collier (2018).
\newblock Which {M}elbourne? augmenting geocoding with maps.

\bibitem[\protect\citeauthoryear{Gritta, Pilehvar, and Collier}{Gritta et~al.}{2019}]{gritta2019pragmatic}
Gritta, M., M.~T. Pilehvar, and N.~Collier (2019).
\newblock A pragmatic guide to geoparsing evaluation.
\newblock {\em Language resources and evaluation\/}, 1--30.

\bibitem[\protect\citeauthoryear{Gritta, Pilehvar, Limsopatham, and Collier}{Gritta et~al.}{2017}]{gritta2017missing}
Gritta, M., M.~T. Pilehvar, N.~Limsopatham, and N.~Collier (2017).
\newblock What's missing in geographical parsing?
\newblock {\em Language Resources and Evaluation\/}, 1--21.

\bibitem[\protect\citeauthoryear{Halterman}{Halterman}{2017}]{halterman2017mordecai}
Halterman, A. (2017, Jan).
\newblock Mordecai: Full text geoparsing and event geocoding.
\newblock {\em The Journal of Open Source Software\/}~{\em 2\/}(9).

\bibitem[\protect\citeauthoryear{Halterman}{Halterman}{2019}]{halterman2019linking}
Halterman, A. (2019).
\newblock Geolocating political events in text.
\newblock In {\em Proceedings of the Third Workshop on Natural Language Processing and Computational Social Science, 17th Annual Conference of the North American Chapter of the Association for Computational Linguistics ({NAACL})}, pp.\  29--39.

\bibitem[\protect\citeauthoryear{Honnibal and Montani}{Honnibal and Montani}{2017}]{honnibal2017spacy}
Honnibal, M. and I.~Montani (2017).
\newblock spacy 2: Natural language understanding with bloom embeddings, convolutional neural networks and incremental parsing.
\newblock {\em To appear\/}.

\bibitem[\protect\citeauthoryear{Kamalloo and Rafiei}{Kamalloo and Rafiei}{2018}]{kamalloo2018coherent}
Kamalloo, E. and D.~Rafiei (2018).
\newblock A coherent unsupervised model for toponym resolution.
\newblock In {\em Proceedings of the 2018 World Wide Web Conference}, pp.\  1287--1296.

\bibitem[\protect\citeauthoryear{Karimzadeh, Pezanowski, MacEachren, and Wallgr{\"u}n}{Karimzadeh et~al.}{2019}]{karimzadeh2019geotxt}
Karimzadeh, M., S.~Pezanowski, A.~M. MacEachren, and J.~O. Wallgr{\"u}n (2019).
\newblock Geotxt: A scalable geoparsing system for unstructured text geolocation.
\newblock {\em Transactions in GIS\/}~{\em 23\/}(1), 118--136.

\bibitem[\protect\citeauthoryear{Leidner}{Leidner}{2008}]{leidner2008toponym}
Leidner, J.~L. (2008).
\newblock {\em Toponym resolution in text: Annotation, evaluation and applications of spatial grounding of place names}.
\newblock Universal-Publishers.

\bibitem[\protect\citeauthoryear{Leidner}{Leidner}{2017}]{leidner2017georeferencing}
Leidner, J.~L. (2017).
\newblock Georeferencing: From texts to maps.
\newblock {\em The International Encyclopedia of Geography\/}.

\bibitem[\protect\citeauthoryear{Lieberman, Samet, and Sankaranarayanan}{Lieberman et~al.}{2010}]{lieberman2010geotagging}
Lieberman, M.~D., H.~Samet, and J.~Sankaranarayanan (2010).
\newblock Geotagging with local lexicons to build indexes for textually-specified spatial data.
\newblock In {\em Data Engineering (ICDE), 2010 IEEE 26th International Conference on}, pp.\  201--212. IEEE.

\bibitem[\protect\citeauthoryear{Radford}{Radford}{2021}]{radford2021regressing}
Radford, B.~J. (2021).
\newblock Regressing location on text for probabilistic geocoding.
\newblock {\em arXiv preprint arXiv:2107.00080\/}.

\bibitem[\protect\citeauthoryear{Speriosu and Baldridge}{Speriosu and Baldridge}{2013}]{speriosu2013text}
Speriosu, M. and J.~Baldridge (2013).
\newblock Text-driven toponym resolution using indirect supervision.
\newblock In {\em ACL}, pp.\  1466--1476.

\bibitem[\protect\citeauthoryear{Wang, Ma, Zheng, Liu, Xie, Li, and Si}{Wang et~al.}{2019}]{wang2019dm_nlp}
Wang, X., C.~Ma, H.~Zheng, C.~Liu, P.~Xie, L.~Li, and L.~Si (2019).
\newblock Dm\_nlp at semeval-2018 task 12: A pipeline system for toponym resolution.
\newblock In {\em Proceedings of the 13th International Workshop on Semantic Evaluation}, pp.\  917--923.

\bibitem[\protect\citeauthoryear{Wick and Boutreux}{Wick and Boutreux}{2011}]{wick2011geonames}
Wick, M. and C.~Boutreux (2011).
\newblock Geonames.
\newblock {\em GeoNames Geographical Database\/}.

\end{thebibliography}

\appendix

%\section{Wiki examples}

%``In Mostar, the ARBiH and HVO forces had clashes in the city and its Bijelo Polje and Raštani suburbs."

\end{document}